\documentclass[runningheads]{llncs}

\usepackage[mobile]{eccv}

\usepackage{eccvabbrv}

\usepackage{graphicx}
\usepackage{booktabs}
\usepackage{tabularx}
\usepackage{multirow}
\usepackage{graphicx}
\usepackage{adjustbox}
\usepackage{pifont}
\usepackage[table]{xcolor}

\usepackage[accsupp]{axessibility}  

\usepackage[pagebackref,breaklinks,colorlinks,citecolor=eccvblue]{hyperref}

\usepackage{orcidlink}
\usepackage{cuted}
\usepackage{tcolorbox}
\newtcolorbox{planbox}[1]{
    colback=gray!5,
    colframe=gray!75,
    title=#1,
    fonttitle=\bfseries
}

\begin{document}

\title{\textit{IAM}: \textcolor{blue}{\textbf{I}}dentity-\textcolor{blue}{\textbf{A}}ware Human \textcolor{blue}{\textbf{M}}otion and Shape Joint Generation}

\titlerunning{IAM}


\author{
Wenqi Jia\inst{1,2} \and
Zekun Li\inst{2,3} \and
Abhay Mittal\inst{2} \and
Chengcheng Tang\inst{2} \and
Chuan Guo\inst{2} \and
Lezi Wang\inst{2} \and
James Matthew Rehg\inst{1} \and
Lingling Tao\inst{2} \and
Sizhe An\inst{2}
}

\authorrunning{Jia et al.}

\institute{
$^{1}$UIUC \quad
$^{2}$Meta Reality Labs \quad
$^{3}$Brown University
}

\maketitle

\begin{center}
    \captionsetup{type=figure}
    \includegraphics[width=0.95\linewidth]{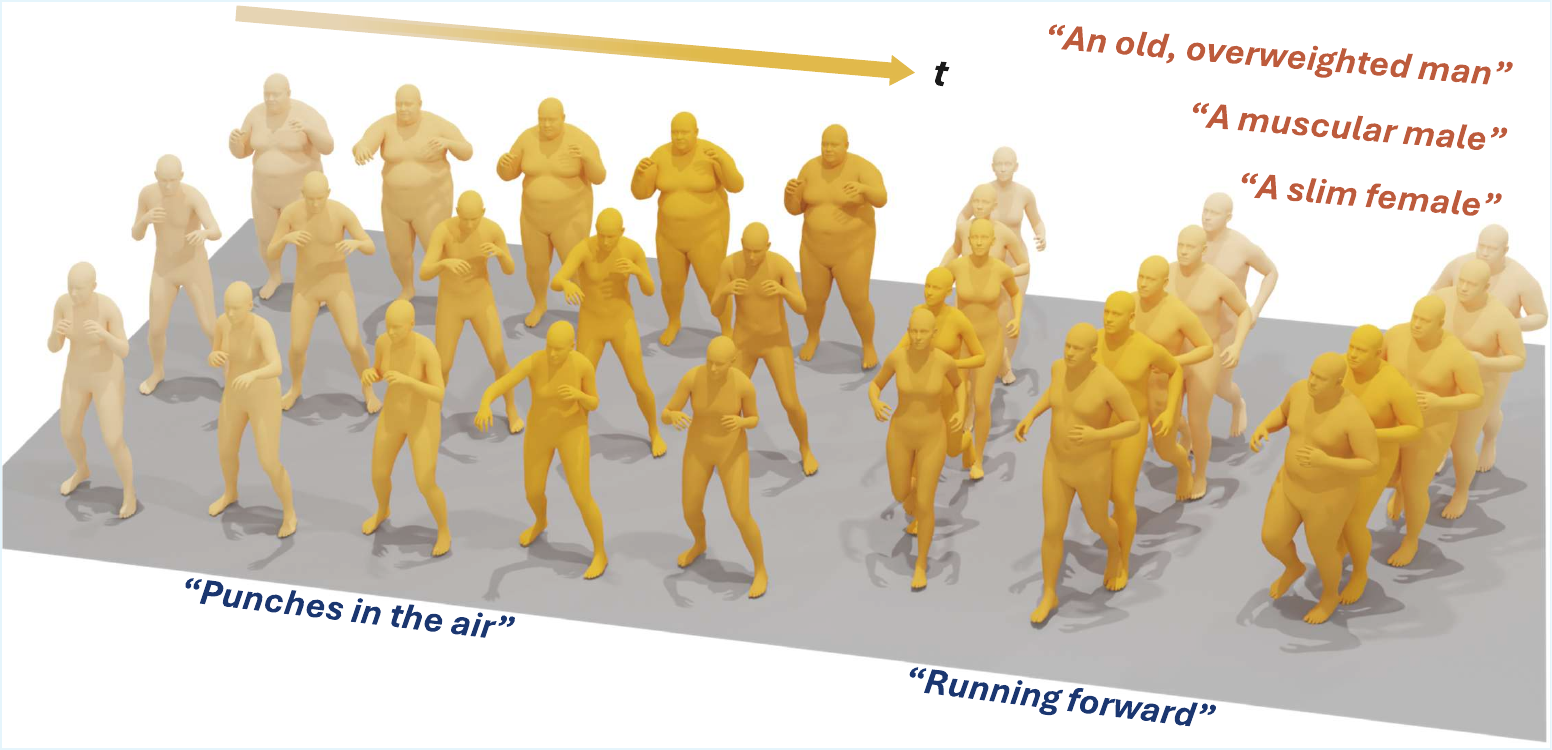}
    \vspace{-10pt}
    \captionof{figure}{\textbf{Identity-Consistent Motion Generation.} 
    Our framework enables decoupled control of action dynamics and subject morphology. Given identity cues and motion prompts, the model synthesizes diverse body shapes while producing motions that remain physically consistent with body morphology.
    }
    \label{fig:teaser}
\end{center}


\newcommand{\wq}[1]{\textcolor{blue}{[wq: #1]}}
\newcommand{\sa}[1]{\textcolor{purple}{[sa: #1]}}
\newcommand{\todo}[1]{\textcolor{red}{[TODO: #1]}}
\newcommand{\kk}[1]{{\color{Emerald} {(ZK: #1)}}}
\newcommand{\lz}[1]{{\color{cyan} {(lezi: #1)}}}
\newcommand{\am}[1]{{\color{purple} {(am: #1)}}}
\newcommand{\cc}[1]{{\color{orange} {(cc: #1)}}}

\begin{abstract}
Recent advances in text-driven human motion generation enable models to synthesize realistic motion sequences from natural language descriptions. However, most existing approaches assume identity-neutral motion and generate movements using a canonical body representation, ignoring the strong influence of body morphology on motion dynamics. In practice, attributes such as body proportions, mass distribution, and age significantly affect how actions are performed, and neglecting this coupling often leads to physically inconsistent motions. We propose an identity-aware motion generation framework that explicitly models the relationship between body morphology and motion dynamics. Instead of relying on explicit geometric measurements, identity is represented using multimodal signals, including natural language descriptions and visual cues. We further introduce a joint motion–shape generation paradigm that simultaneously synthesizes motion sequences and body shape parameters, allowing identity cues to directly modulate motion dynamics. Extensive experiments on motion capture datasets and large-scale in-the-wild videos demonstrate improved motion realism and motion–identity consistency while maintaining high motion quality. 
Project Page: \href{https://vjwq.github.io/IAM}{https://vjwq.github.io/IAM}.
\end{abstract}

\section{Introduction}
\label{sec:intro}
Human motion generation from natural language has made rapid progress in recent years, enabling models to synthesize diverse and realistic action sequences from textual descriptions~\cite{guo2024momask,tevet2022human,jiang2024motiongpt,zhang2023generating,xiao2025motionstreamer,li2026llamo,fan2025go}. Recent advances in diffusion models, transformer architectures, and discrete motion tokenization have significantly improved motion quality and semantic alignment, making text-to-motion generation a promising paradigm for character animation~\cite{zhang2023tapmo}, digital avatars~\cite{jiang2025solami,zhang2025vibes}, and embodied simulation~\cite{wu2025uniphys,xie2026textop}.

Despite these advances, most existing approaches share an implicit assumption: motion dynamics are universal and independent of the person performing them. In practice, however, human motion is inseparable from the physical identity of the body that produces it. Attributes including height, limb proportions, and mass distribution directly modulate how movements are executed. For example, two individuals performing the same ``jogging'' instruction exhibit distinct stride lengths, joint trajectories, and temporal dynamics dictated by their specific musculoskeletal constraints. These morphological factors shape coordination patterns, forming characteristic motion signatures intrinsically tied to an individual's physical identity~\cite{jiang2019synthesis}.

Nevertheless, most existing human motion generation models represent movement using a canonical, standardized skeleton and body shape template~\cite{SMPL:2015,SMPL-X:2019}. Under this paradigm, identity attributes are either ignored during synthesis or introduced post-hoc through retargeting~\cite{tripathi2024humos,zhang2023skinned} or skeleton rescaling~\cite{Guo_2022_CVPR}. While computationally convenient, this decoupled design treats identity as a superficial visual attribute rather than a fundamental structural prior that governs motion dynamics. Consequently, generated motions often reflect ``averaged'' dynamics. 
When transferred to characters with diverse body shapes, these generic trajectories may lead to visual artifacts or unnatural coordination, hindering the deployment in downstream tasks that demand high individual fidelity, such as personalized avatar animation~\cite{zhang2024generative, shi2025motionpersona} and physically-informed embodied simulation~\cite{Lee:2021:Parameterized, xu2023adaptnet}.

Recent works have begun exploring shape-conditioned motion synthesis by incorporating body morphology into the generation process~\cite{liao2025shape,tripathi2024humos}. For instance, Shape My Moves~\cite{liao2025shape} utilizes explicit anthropometric measurements alongside motion prompts. However, such approaches typically rely on precise numerical descriptors that are difficult for average users to provide. More fundamentally, these methods treat body morphology merely as an external conditioning signal, leaving the underlying motion engine identity-agnostic. This formulation fails to capture the \textit{intrinsic coupling} between form and function; the model learns to ``warp'' a universal motion to fit a shape, rather than learning how a specific body type inherently moves.

In this work, we propose an \textit{Identity-Aware Motion Generation} framework that explicitly models the interdependence between body morphology and motion dynamics. Instead of relying on rigid geometric measurements, we represent human identity through multimodal signals, including natural language descriptions and visual observations. Textual descriptions convey high-level semantic attributes (e.g., ``a tall, athletic male''), while images provide fine-grained structural cues such as proportions and mass distribution that are difficult to articulate in prose.

To achieve morphologically grounded synthesis, we introduce a \textit{joint motion--shape generation paradigm}. By modeling the joint distribution of motion sequences and body parameters, identity cues directly inform the generation process rather than acting as a post-hoc constraint. This approach ensures that synthesized trajectories are inherently consistent with the character's physical frame. Our framework is model-agnostic and can be integrated into both diffusion-based and token-based architectures. We evaluate our framework using motion capture data and large-scale in-the-wild videos, supported by a scalable pipeline that extracts motion, shape, and multimodal identity annotations. Experimental results demonstrate that explicitly modeling identity-motion interdependence significantly improves morphological consistency and motion realism while maintaining high generation quality, as illustrated in Fig.~\ref{fig:teaser}.


Our contributions are summarized as follows:
\begin{itemize}
    \item \textbf{Joint Motion--Shape Paradigm:} We propose \textit{IAM}, a novel framework that learns the joint distribution of motion and body morphology, ensuring synthesized dynamics are inherently grounded in physical identity.
    \item \textbf{Multimodal Identity Prior:} We leverage synergistic text and visual cues as identity priors, providing a flexible interface that captures holistic body attributes from natural language and images without relying on manual geometric descriptors.
    \item \textbf{Universal Compatibility and Efficacy:} We demonstrate our identity-aware design is model-agnostic across discrete and continuous backbones. Notably, \textit{IAM} significantly boosts motion quality (FID) and identity consistency ($\beta$ Dist.) on diffusion architectures, achieving state-of-the-art results.
\end{itemize}

\section{Related Works}
\label{sec:related}


\subsection{Text-to-Motion Synthesis}

Text-to-motion (T2M) synthesis~\cite{guo2024momask,xiao2025motionstreamer,zhang2022motiondiffuse,jiang2023motiongpt,chen2025language,wen2025hy,chen2023executing,zhong2023attt2m} aims to generate human motion sequences from natural language descriptions, typically represented in high-dimensional pose parameterizations such as 263-dim~\cite{Guo_2022_CVPR} or 272-dim formats~\cite{xiao2025motionstreamer,lu2025scamo,fan2025go,li2026llamo}. Recent approaches have explored different generative paradigms. Diffusion-based methods~\cite{tevet2022human,chen2023executing,zhang2022motiondiffuse} achieve strong motion quality and diversity through iterative denoising. Transformer-based methods, such as MoMask~\cite{guo2024momask}, leverage masked modeling to improve temporal coherence. Concurrently, token-based autoregressive models, including T2M-GPT~\cite{zhang2023generating} and Motion-GPT~\cite{jiang2024motiongpt}, discretize motion into token sequences, enabling long-range dependency modeling via language modeling techniques. Despite these advances, most T2M methods assume a fixed canonical body representation and generate motion independent of subject identity or morphology. As a result, the influence of individual body proportions, age, or gender on motion dynamics is largely ignored, limiting the realism and identity consistency of synthesized motions.

\subsection{Human Shape-Conditioned Motion Synthesis}
Recent works begin to explore incorporating identity or body shape into motion generation~\cite{bjorkstrand2025unconditional,wang2025generating,kim2025personabooth,liao2025shape,tripathi2024humos}. Shape My Moves~\cite{liao2025shape} models body shape jointly with motion, but relies on accurate numerical conditioning, limiting flexibility and robustness in identity transfer. HUMOS~\cite{tripathi2024humos} improves identity controllability through identity-aware conditioning; however, it focuses primarily on motion retargeting and does not enable joint synthesis of motion and shape from text. Other works focus on stylized motion generation.  SMooDi~\cite{zhong2024smoodi} and LoRA-MDM~\cite{sawdayee2025dance} enable stylistic control over generated motion by conditioning on expressive attributes. However, these approaches primarily model stylistic variations rather than physical morphology. Consequently, the effects of intrinsic body attributes, such as body shape, gender, or age, remain entangled with stylistic factors. In contrast, our work explicitly models body shape-aware conditioning and enables joint synthesis of identity-consistent body shape and physically plausible motion directly from text.

\section{Method}
\label{sec:method}
\begin{figure}[tb]
  \centering
    \includegraphics[width=\linewidth]{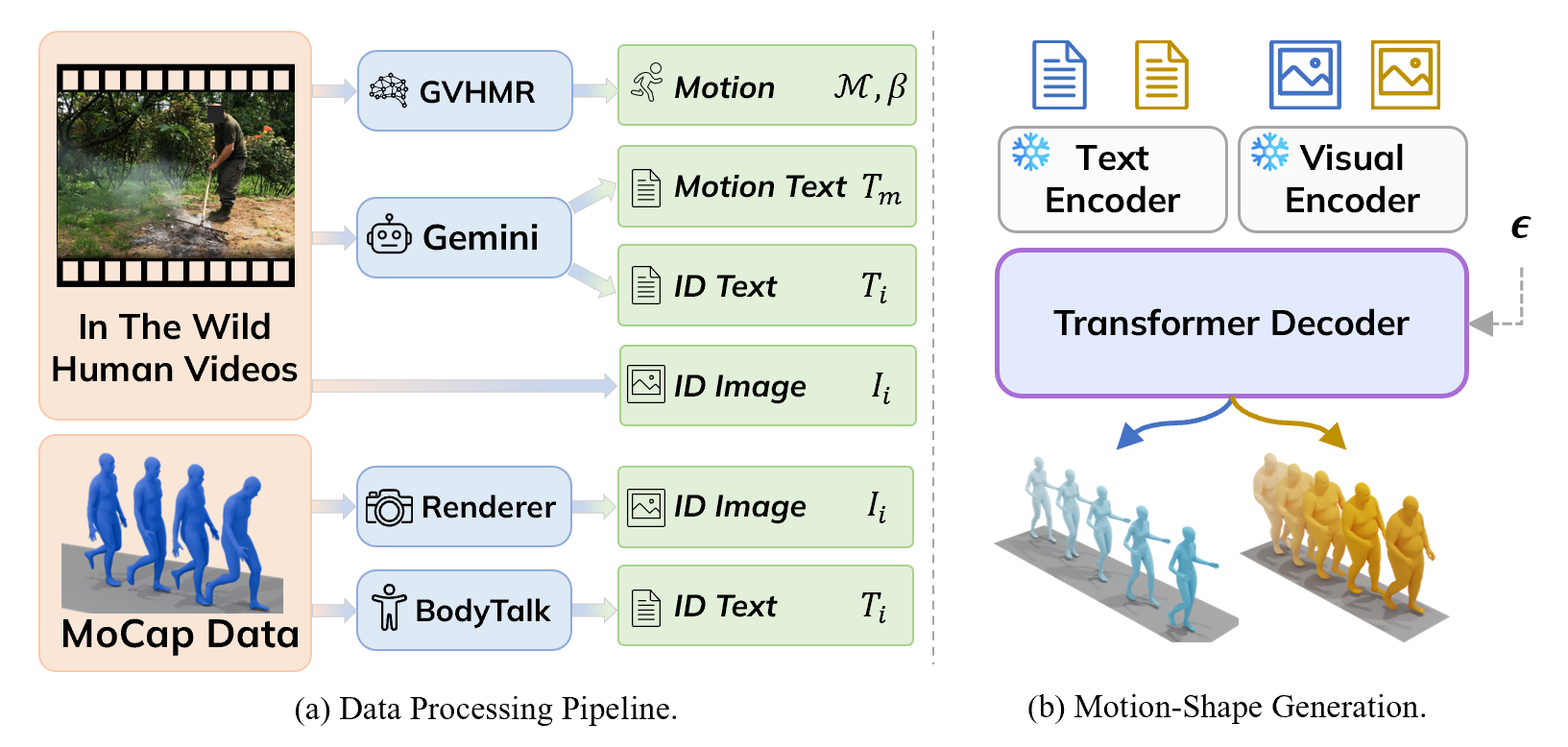}
    \caption{\textbf{Overview of the proposed framework.} (a) \textbf{Data Processing Pipeline}: We extract motion sequences $\mathcal{M}$, shape parameters $\beta$, and multimodal identity descriptions ($T_i, I_i$) from diverse sources including in-the-wild videos and MoCap data. (b) \textbf{Motion-Shape Generation}: A multimodal identity conditioning framework integrates textual and visual priors through frozen encoders to jointly generate identity-consistent motion sequences and body shapes via a diffusion model.
    }
    \vspace{-10pt}
    \label{fig:methodfig}
\end{figure}

We aim to synthesize identity-consistent human motion from textual motion descriptions under multimodal identity conditioning while preserving semantic plausibility and structural realism. Fig.~\ref{fig:methodfig} illustrates the overview architecture of the motion and shape joint generation framework. We present the task formulation (Sec.~\ref{subsec:task_formulation}), data processing pipeline (Sec.~\ref{sec:dataprocess}), multimodal identity conditioning mechanism (Sec.~\ref{subsection:conditioning}), and joint motion–shape generation paradigm (Sec.~\ref{subsec:paradigm}).
\subsection{Task Formulation}
\label{subsec:task_formulation}
Formally, we define the motion prompt as a natural language description of the action, denoted as $T_m$, and the identity condition as a multi-modal representation consisting of a semantic identity description $T_i$ and an optional visual prior $I_i$. Together, the identity condition is represented as $\mathcal{C}_i = \{T_i, I_i\}$. 
Given the motion prompt $T_m$ and identity condition $\mathcal{C}_i$, the objective is to generate: (1) a motion sequence $\mathcal{M} = \{x_1, x_2, \dots, x_L\}$, where each frame $x_t \in \mathbb{R}^{272}$ represents the pose and motion features at time $t$~\cite{xiao2025motionstreamer}, and (2) the corresponding body shape parameters $\beta \in \mathbb{R}^{10}$, describing the subject's morphology. 

\subsection{Identity Representation}
\label{sec:dataprocess}
We represent human identity using two complementary modalities: text-based semantic descriptions and image-based structural priors. We consider two datasets with different annotation settings: the HumanML3D benchmark with ground-truth SMPL parameters and our large-scale in-the-wild IdentityMotion dataset with multimodal annotations, described in section \ref{sec:experiments}. We employ GVHMR~\cite{shen2024gvhmr} to reconstruct pseudo ground-truth motion sequences $\mathcal{M}$ and SMPL-X body shape parameters $\beta$ for the in-the-wild videos, which serve as supervision signals.
\vspace{-10pt}
\subsubsection{Text-based Identity Representation.}
We define the semantic identity $T_i$ as a natural language description focusing on the subject's physique and morphological attributes. We employ two distinct pipelines to ensure $T_i$ is grounded in body shape reality:
\begin{itemize}
    \item \textbf{Knowledge-based Synthesis:} For the \textit{HumanML3D} benchmark, we retrieve the ground-truth body shape parameters $\beta \in \mathbb{R}^{10}$ from the source AMASS collection~\cite{AMASS:ICCV:2019}. These numerical parameters are mapped to anatomical keywords via the \textit{Body Talk} framework~\cite{streuber2016body}. We then utilize Llama 3.2~\cite{grattafiori2024llama} to refine these keywords into fluent, natural body descriptions.
    \item \textbf{VLM-based Annotation:} For our large-scale in-the-wild dataset \textit{IdentityMotion}, we leverage Gemini 2.5 Pro~\cite{comanici2025gemini} to perform multimodal annotation. Following~\cite{fan2025zerozeroshotmotiongeneration}, we carefully design structured prompts to decouple motion dynamics $T_m$ from identity attributes $T_i$, yielding consistent and descriptive identity tokens for over 200k sequences. We further employ Llama 3.2~\cite{grattafiori2024llama} to rewrite $T_m$, ensuring that it contains no identity-related information.
\end{itemize}

\subsubsection{Image-based Identity Representation}
To provide precise structural and appearance guidance, we complement textual descriptions with visual priors $I_i$. This modality captures fine-grained structural cues like limb proportions and torso-to-leg ratios, which is often elude concise textual description.
\begin{itemize}
    \item \textbf{Synthesized Priors:} Since the original video data for \textit{HumanML3D} is unavailable, we reconstruct identity references by rendering SMPL meshes. Specifically, we utilize retrieved $\beta$ parameters to generate shaded images in a canonical front-view, providing a normalized geometric reference for identity conditioning.
    \item \textbf{Authentic Priors:} For our \textit{IdentityMotion} dataset, we leverage the high-fidelity visual information inherent in the source videos. We extract a representative keyframe from each sequence to serve as the identity reference $I_i$. These real-world images provide authentic appearance priors, capturing nuanced physical characteristics and surface details that significantly enhance the diversity and realism of the identity-conditioned generation.
\end{itemize}
\subsection{Multimodal Identity Conditioning Mechanism}
\label{subsection:conditioning}
We design a multimodal conditioning mechanism that integrates motion prompts $T_m$ with identity conditions $\mathcal{C}_i = \{T_i, I_i\}$. As shown in Fig.~\ref{fig:methodfig} (b), the conditioning signals and strategy are described as follows.\\
\textbf{Textual Encoding.} We employ a frozen text encoder to extract semantic embeddings from the motion description $T_m$ and the identity description $T_i$. These textual inputs are encoded into a latent representation $E_{txt} \in \mathbb{R}^{L \times d_{txt}}$, where $L$ denotes the sequence length and $d_{txt}=768$. A learnable projection layer is subsequently applied to map these embeddings into the hidden dimension $d=512$. \\
\textbf{Visual Encoding.} For the visual prior $I_i$, a frozen image encoder extracts a 512-dimensional feature vector, capturing identity-related attributes such as age, gender, and body morphology, as well as fine-grained structural cues like limb proportions. This image feature is projected into a visual embedding $E_{img} \in \mathbb{R}^{1 \times d}$ to maintain dimensional consistency with the textual features. \\
\textbf{Fusion Strategy.} We adopt a late fusion strategy by treating the visual and textual priors as distinct tokens within a unified condition sequence. Specifically, the final conditional representation $C$ is constructed by concatenating the projected textual tokens and the visual token along the sequence dimension: $C = [E_{txt}; E_{img}] \in \mathbb{R}^{(L+1) \times d}$. To support classifier-free guidance \cite{ho2022classifier}, we include a joint dropout mechanism during training, where both textual and visual embeddings are replaced by a learnable null token with a probability of 10\%. This enables the model to effectively navigate between conditional and unconditional score estimates during the inference stage.

\subsection{Joint Motion-Shape Generative Paradigms}
\label{subsec:paradigm}
Our framework is model-agnostic and can be integrated into various generative architectures to perform joint synthesis of motion sequences $\mathcal{M}$ and body shape parameters $\beta \in \mathbb{R}^{10}$. The core principle is to reformulate the standard motion generation task into a joint density estimation problem $p(\mathcal{M}, \beta | \mathcal{C}_i)$, ensuring the generated dynamics are inherently grounded in the synthesized morphology.

\subsubsection{Diffusion-based Paradigm.}
Building upon~\cite{tevet2024closd}, we adapt the denoising network to predict the joint distribution of $x_t$ (pose) and $\beta$ (shape), as illustrated in Fig.~\ref{fig:methodfig}. Instead of treating motion and body shape as decoupled tasks, we propose a \textit{unified joint representation} by augmenting the motion feature space. Specifically, we concatenate the 10-dimensional shape parameters $\beta$ to the motion representation $x \in \mathbb{R}^{272}$ at each frame, forming a joint state vector $z = [x; \beta] \in \mathbb{R}^{282}$. The denoising network $\epsilon_\theta$ is then trained to reconstruct this joint manifold from Gaussian noise, guided by the multimodal identity condition $\mathcal{C}_i$. The training objective is defined as a holistic Mean Squared Error (MSE) loss over the joint space:
\begin{equation}
    \mathcal{L}_{joint} = \mathbb{E}_{t, z_0, \epsilon} \left[ \| \epsilon - \epsilon_\theta(z_t, t, \mathcal{C}_i) \|^2 \right]
\end{equation}
This formulation forces the model to capture the intrinsic correlations between temporal dynamics and static morphology. By backpropagating through a single unified loss, the diffusion process naturally learns to maintain identity consistency throughout the generated sequence.

\subsubsection{VQ-based Paradigm.}
For the VQ-based paradigm, we adapt the MoMask framework by incorporating an auxiliary body shape prediction head into the generative Transformer. While the motion is discretized into tokens through a frozen residual vector quantizer (RVQ), the Transformer is tasked with learning the joint distribution of these discrete tokens and the continuous shape parameters $\beta$. Specifically, the Transformer output is bifurcated into a classification head for motion tokens and a regression head for body shape. Given the identity condition $\mathcal{C}_i$, the model predicts both the token logits and the shape parameters $\hat{\beta}$. The total training objective is formulated as a multi-task loss:
\begin{equation}
    \mathcal{L}_{VQ} = \mathcal{L}_{ce} + \gamma \| \beta - \hat{\beta} \|^2
\end{equation}
where $\mathcal{L}_{ce}$ is the cross-entropy loss for masked token prediction, and the second term is the Mean Squared Error (MSE) for shape regression. The hyperparameter $\gamma$ was set to 0.1, serves as a balancing coefficient to align the gradients of the discrete and continuous optimization targets. Notably, we keep the original RVQ frozen to leverage its high-quality motion priors while enabling identity-specific morphological generation.

\begin{figure}[tb]
  \centering
  \begin{subfigure}{0.495\linewidth}
    \includegraphics[width=\linewidth]{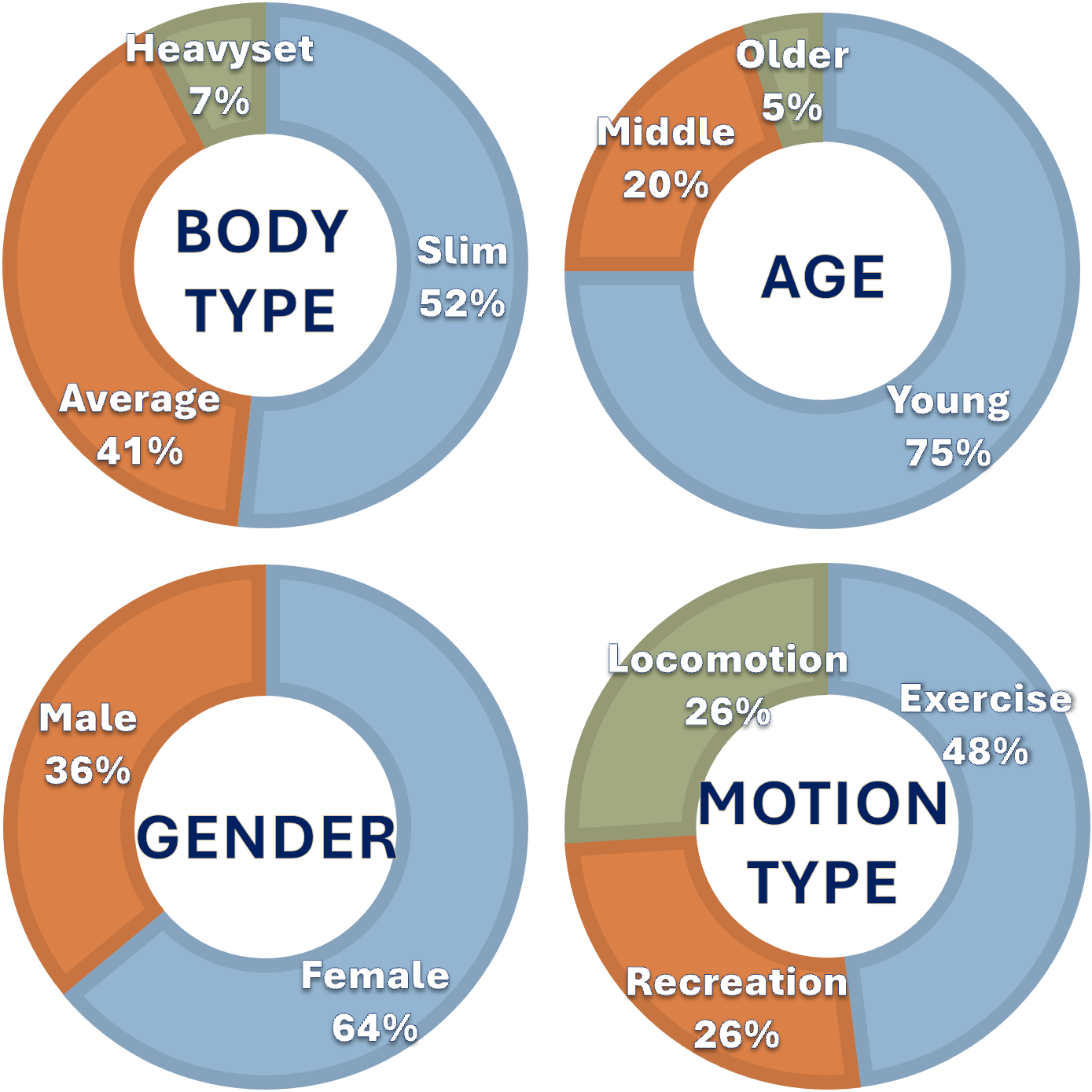}
    \label{fig:short-a}
  \end{subfigure}
  \hfill
  \begin{subfigure}{0.495\linewidth}
    \includegraphics[width=\linewidth]{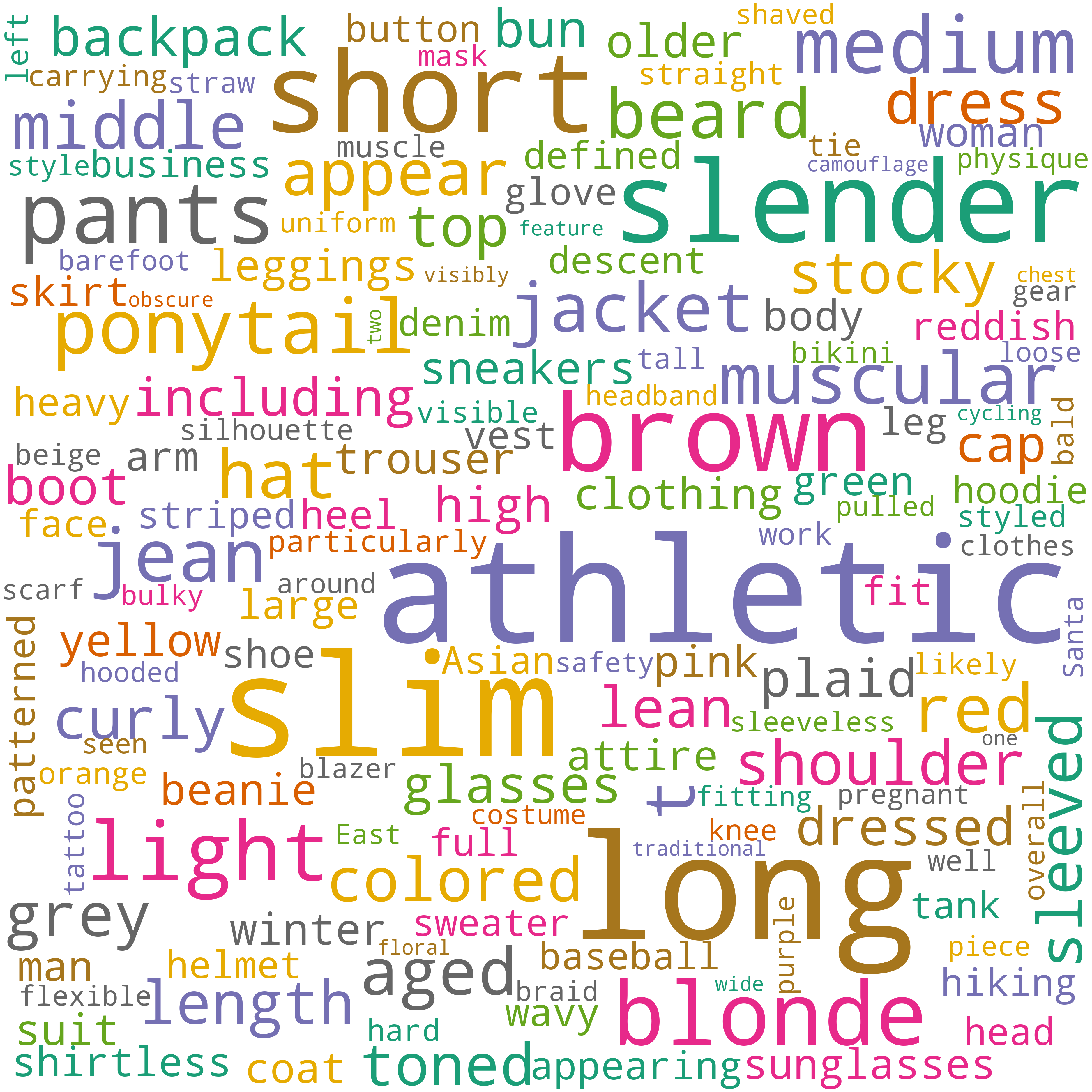}
    \label{fig:short-b}
  \end{subfigure}
  \caption{\textbf{Overview of IdentityMotion Dataset.} Left: Distribution of identity and motion attributes, including body type, age, gender, and motion category. Right: Word cloud of identity descriptions, highlighting common appearance attributes.}
  \label{fig:wordcloud}
\end{figure}

\newcommand{\neut}{\textcolor{OrangeRed}{\bullet}\xspace}          
\newcommand{\iden}{\textcolor{ForestGreen}{\bullet}\xspace}        
\newcommand{\imgf}{\textcolor{RoyalBlue}{\bullet}\xspace}          

\newcommand{\cmark}{\checkmark}
\newcommand{\xmark}{\ding{55}}

\section{Experiments}
\label{sec:experiments}
We conduct extensive experiments to evaluate three key aspects: (1) motion generation quality, (2) identity–shape consistency, and (3) generalization to unseen identities. Our evaluation examines motion quality, text alignment, body-shape reconstruction accuracy, and generalization to unseen identities. Comprehensive quantitative (Sec.~\ref{sec:quanti}) and qualitative results (Sec.~\ref{sec:quali}) demonstrate the effectiveness of the proposed paradigm.
\vspace{-10pt}
\subsection{Datasets}
\noindent \textbf{HumanML3D.} We conduct evaluations on HumanML3D, which contains 14,616 motions and 44,970 text descriptions. To enable identity-aware generation, we augment this dataset with ground-truth SMPL shape parameters $\beta \in \mathbb{R}^{10}$ retrieved from the AMASS~\cite{AMASS:ICCV:2019} collection. The augmented dataset covers 449 unique identities, including 263 males and 186 females, with a diverse body type distribution of 116 slim, 269 average, and 64 heavyset individuals. As detailed in Sec.~\ref{sec:dataprocess}, body shapes are converted into natural language identity descriptions and synthesized structural priors to serve as multi-modal conditioning signals. We use the standard split to evaluate the model's performance in joint motion-shape synthesis. 

\noindent \textbf{IdentityMotion.} To improve identity diversity during training, we further curate a large-scale in-the-wild corpus containing over 200k motion sequences via a VLM-based pipeline. Unlike the synthesized priors in HumanML3D, \textit{IdentityMotion} provides authentic appearance priors extracted directly from source videos, capturing nuanced human characteristics. The dataset features a balanced distribution across various attributes, including body type (52\% slim, 41\% average, 7\% heavyset), gender (64\% female, 36\% male), and age (see Fig.~\ref{fig:wordcloud}). We use IdentityMotion for both training and evaluation, and assess zero-shot generalization by testing on subjects strictly disjoint from those seen during training.

\subsection{Implementation Details}
Our model is built upon a Transformer-based architecture with 8 layers, 4 attention heads, and a latent dimension of 256. We follow~\cite{tevet2024closd} to use a frozen DistilBERT~\cite{sanh2019distilbert} as text encoder, and have a frozen CLIP~\cite{radford2021learning} as image feature encoder. We train our model on HumanML3D for 400k steps using 2 NVIDIA H100 GPUs with a batch size of 64 per GPU and a learning rate of $1 \times 10^{-4}$. For IdentityMotion, we train for 200k steps using 4 NVIDIA H100 GPUs with a batch size of 256 per GPU and a learning rate of $2 \times 10^{-4}$. All experiments use the AdamW optimizer with GELU activations. 

\subsection{Evaluation Metrics}
\noindent \textbf{Text-to-Motion.} We follow standard T2M protocols~\cite{Guo_2022_CVPR}, reporting \textit{Fréchet Inception Distance (FID) $\downarrow$}, \textit{R-Precision (R@k, $k \in \{1,2,3\}$) $\uparrow$}, \textit{Multi-Modal Distance (MM-D) $\downarrow$}, and \textit{Diversity (Div) $\uparrow$} to evaluate motion quality, text alignment, and motion variation.

\noindent \textbf{Shape Metrics.} We evaluate body shape accuracy in both parameter and geometry space. Following~\cite{choutas2022accurate}, we report \textit{Body Measurement error $\downarrow$} (Height, Chest, Waist, Hips) alongside SMPL/SMPL-X $\beta$ parameter reconstruction (\textit{$\beta$-Dist $\downarrow$}). For HumanML3D and IdentityMotion, we utilize the first 10 dimensions of SMPL and SMPL-X $\beta$ respectively to maintain a consistent scale. For geometric fidelity, we report \textit{V2V error $\downarrow$} and \textit{P2P$_{20k}$ error $\downarrow$} (20K sampled surface points). Note that \textit{$\beta$-Dist} for "Real motion" denotes the average $L_2$ distance between individual ground-truth $\beta$ and the dataset mean, reflecting inherent shape variance.

\subsection{Baselines}
We train and evaluate our diffusion-based model and the VQ-based baseline on HumanML3D. For IdentityMotion, we evaluate the diffusion-based model on held-out identities to assess zero-shot identity generalization, and provide a qualitative comparison with~\cite{liao2025shape}. Detailed quantitative and visual results are presented in~\ref{sec:quanti} and~\ref{sec:quali}.

\noindent \textbf{Diffusion-Based Model.} 
We train an improved version of MDM~\cite{tevet2024closd} under different conditioning configurations, including identity-neutral motion prompts, identity-aware motion prompts, and image features, denoted as $\neut$, $\iden$, and $\imgf$, respectively. This setup allows us to systematically analyze the contribution of each modality to body shape accuracy and motion fidelity.

\noindent \textbf{VQ-Based Model.} 
We adapt the MoMask~\cite{guo2024momask} framework by incorporating an auxiliary body shape prediction head into the generative Transformer. The original RVQ codebook is kept frozen, and only the Transformer is trained to model the joint distribution of motion sequences and body identities.

\noindent \textbf{Shape My Moves}~\cite{liao2025shape} employs templated numeric body measurements for identity-aware motion generation. As its training code and optimization pipeline are not publicly released, we do not retrain the model and instead provide a qualitative comparison on the HumanML3D and IdentityMotion test sets.








\begin{table*}[t]
\caption{
\textbf{Quantitative Results on HumanML3D.}
    Our model supports joint motion and body shape generation under multiple conditioning signals,
    including: \\ \protect$\neut$ \textcolor{OrangeRed}{Identity-neutral motion prompts}, 
    \protect$\iden$ \textcolor{ForestGreen}{Identity-aware motion prompts}, 
    \protect$\imgf$ \textcolor{RoyalBlue}{Image features}.
    Colored dots in the \textit{Cond.} column indicate the inputs used by each method.
}
\label{tab:results_h3d}
\centering
\small
\begin{tabular}{l c c cccc cc c}
\toprule
\multirow{2}{*}{Methods} & \multirow{2}{*}{Cond.} & \multirow{2}{*}{$\beta$ Gen.} & \multicolumn{6}{c}{Text-to-Motion Metrics} & \multirow{2}{*}{\shortstack{$\beta$ Dist.\\$\downarrow$}} \\
\cmidrule(lr){4-9}
& & & FID $\downarrow$ & R@1 $\uparrow$ & R@2 $\uparrow$ & R@3 $\uparrow$ & MM-D $\downarrow$ & Div $\rightarrow$ & \\
\midrule
Real motion & -- & -- & 0.002 & 0.702 & 0.864 & 0.914 & 15.151 & 27.492 & 3.270 \\
\midrule

VQ Based & $\neut$ & \xmark & 10.952 & 0.625 & 0.801 & 0.871& 15.690 & 27.261 & / \\
VQ Based & $\iden$ & \cmark & 11.34 & 0.585  & 0.765  & 0.838 & 16.425  & 27.562  & 1.359 \\

\midrule
Diffusion Based & $\neut$ & \xmark & 13.207 & \textbf{0.701} & \textbf{0.831} & \underline{0.882} & \textbf{15.285} & 27.292 & / \\
Diffusion Based & $\iden$ & \cmark & \underline{7.395} & \underline{0.659} & \underline{0.820} & \textbf{0.884} & \underline{15.537} & 27.133 & 1.190 \\
Diffusion Based & $\neut \imgf$ & \cmark & \textbf{7.371} & 0.642 & 0.812 & 0.873 & 15.685 & 27.142 & \textbf{0.647} \\
Diffusion Based & $\iden \imgf$ & \cmark & 7.774 & 0.640 & 0.812 & 0.871 & 15.806 & 27.121 & \underline{0.711} \\

\bottomrule
\end{tabular}
\end{table*}

\subsection{Quantitative Results}
\label{sec:quanti}

\noindent \textbf{Performance on HumanML3D.} Table~\ref{tab:results_h3d} summarizes the performance on the HumanML3D dataset. Compared to the VQ-based baseline, the diffusion-based variants consistently achieve superior results in both motion quality (FID) and body shape accuracy ($\beta$ Dist.). Specifically, our diffusion model conditioned on both identity-aware prompts and image features ($\imgf\iden$) achieves an FID of \textbf{7.371} and the lowest $\beta$ Dist. of \textbf{0.647}. This demonstrates that the diffusion framework better captures the complex joint distribution of dynamic motions and static body structures.

\noindent \textbf{Effect of Conditioning Signals.} The results in Table~\ref{tab:results_h3d} and Table~\ref{tab:body_meas} highlight the importance of multi-modal identity conditioning. While models using only identity-neutral prompts ($\neut$) fail to generate diverse body shapes, the inclusion of identity-aware text ($\iden$) significantly reduces the $\beta$ Dist.. The best performance is reached by incorporating image features ($\imgf$), which provide fine-grained geometric priors. As shown in Table~\ref{tab:body_meas}, the variant with dual conditioning ($\imgf\iden$) on the HumanML3D dataset reaches a reconstruction accuracy comparable to ShapeMyMoves, with a height error of only 5.8 mm. 

\noindent \textbf{Paradigm Insights.} While our framework is inherently model-agnostic, the performance gains vary across generative paradigms. In particular, the improvements are more pronounced for diffusion-based models, likely due to their stronger capacity for modeling complex joint distributions of motion dynamics and body geometry. In contrast, while capable of synthesizing accurate human shape, VQ-based models introduces a mild trade-off in motion quality after injecting identity-aware constraints, suggesting that diffusion architectures are better suited for jointly modeling identity and motion.

\noindent \textbf{Generalization on IdentityMotion.} To evaluate the generalization capabilities on unseen subjects, we retrain and test the diffusion-based variants on the IdentityMotion dataset. As shown in Table~\ref{tab:results_sstk}, despite the challenges of a zero-shot setting with unseen identities, our model maintains competitive motion fidelity. Notably, the dual-conditioned model ($\imgf\iden$) achieves both the lowest FID score of \textbf{23.174} and shape preservation with a $\beta$ Dist. of \textbf{1.279}, significantly outperforming the single-condition variants. This validates that our model does not merely memorize training identities but learns to map multi-modal identity descriptors to the underlying human body shape space. Furthermore, the superior performance of the dual-conditioned model suggests that, in large-scale real-world settings with diverse human identities, capturing realistic motion dynamics requires combining complementary identity cues from both textual descriptions and visual appearance. This contrasts with smaller-scale datasets such as HumanML3D, where the limited diversity of performers makes single modality largely sufficient for identity conditioning.
The shape reconstruction accuracy reported in Table~\ref{tab:body_meas} is comparable to the state-of-the-art shape estimation performance of Shapy~\cite{choutas2022accurate} on unseen images. This indicates that our model achieves a physically plausible shape estimation range while uniquely supporting the additional task of identity-aware motion generation.

\begin{table*}[t]
\caption{
\textbf{Zero-shot Quantitative Results on IdentityMotion.} We retrain and evaluate the diffusion-based variants on the IdentityMotion dataset. The results are reported on a test set with unseen identities from the training set to evaluate shape-motion generation capability.
}
\label{tab:results_sstk}
\centering
\small
\begin{tabular}{l c c cccc cc c}
\toprule
\multirow{2}{*}{Methods} & \multirow{2}{*}{Cond.} & \multirow{2}{*}{$\beta$ Gen.} & \multicolumn{6}{c}{Text-to-Motion Metrics} & \multirow{2}{*}{\shortstack{$\beta$ Dist.\\$\downarrow$}} \\
\cmidrule(lr){4-9}
& & & FID $\downarrow$ & R@1 $\uparrow$ & R@2 $\uparrow$ & R@3 $\uparrow$ & MM-D $\downarrow$ & Div $\rightarrow$ & \\
\midrule
Real motion & -- & -- & 0.002 & 0.565 & 0.822 & 0.916 & 0.750 & 30.2919 & 2.972 \\
\midrule
Diffusion Based & $\neut$ & \xmark & 31.799 & \underline{0.629} & \underline{0.847}  & \underline{0.927}  & \textbf{0.781}  & 29.718  & / \\
Diffusion Based & $\iden$ & \cmark &  \underline{28.667} & \textbf{0.641} & \textbf{0.857} & \textbf{0.935}  & 0.754 & 29.879 & 1.452 \\
Diffusion Based & $\neut \imgf$ & \cmark & 31.726 & 0.606 & 0.834 & 0.918 & \underline{0.778} & 29.866 & \underline{1.392} \\
Diffusion Based & $\iden \imgf$ & \cmark & \textbf{23.174} & 0.609 & 0.836 & 0.926 & 0.771 & 30.032 & \textbf{1.279} \\
\bottomrule
\end{tabular}
\end{table*}

\begin{table}[t]
\centering
\small
\setlength{\tabcolsep}{4pt}
\caption{\textbf{Body Shape Reconstruction Accuracy in mm.} Our diffusion-based model reaches comparable performance with ShapeMyMoves.}
\label{tab:body_meas}
\begin{tabular}{llccccccc}
\toprule
Method & Cond.& Dataset & Height & Chest & Waist & Hips & $\mathrm{V2V}$ & $\mathrm{P2P}_{20k}$ \\ 
\midrule
ShapeMyMoves & -- & HumanML3D & 5.8 & 6.9 & 10.6 & 6.0 & / & / \\
\midrule
Diffusion Based & $\iden$ & HumanML3D & 18.8 & 16.9 & 23.7 & 16.7 & 7.9 & 7.0 \\
Diffusion Based & $\iden \imgf$ & HumanML3D & 5.8 & 8.6 & 12.6 & 6.8 & 3.0 & 2.8 \\
Diffusion Based & $\iden \imgf$ & IdentityMotion & 53.5 & 67.9 & 77.3 & 32.1 & 17.6 & 19.1 \\
\bottomrule
\end{tabular}
\end{table}

\vspace{-3pt}
\subsection{Qualitative Results}
\label{sec:quali}
\begin{figure}[h]
  \centering
  \includegraphics[width=\linewidth]{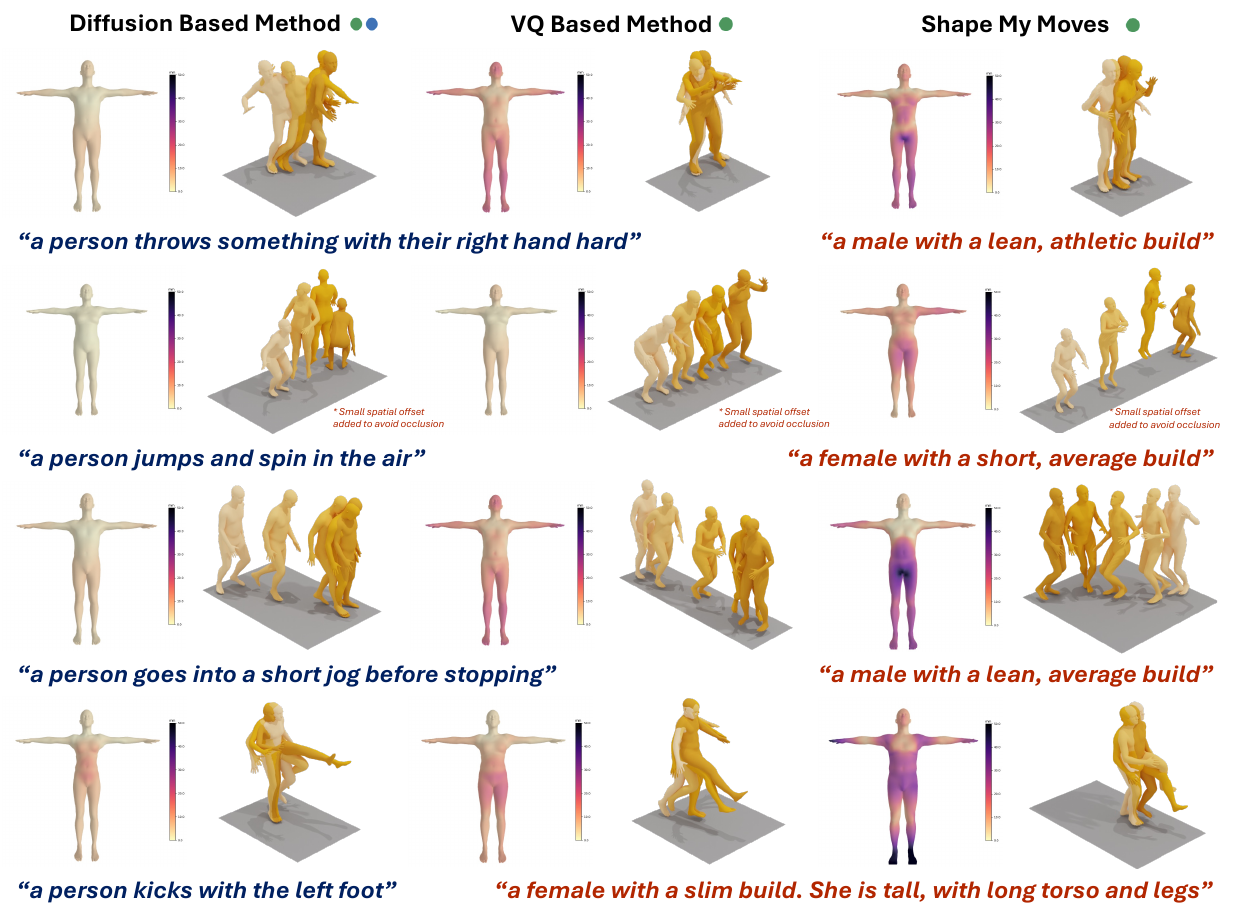}
  \caption{
    \textbf{Qualitative Comparison on the HumanML3D Test Set.}
    Given the same action prompt and identity description, different methods generate motion sequences conditioned on the same inputs. 
    The colored meshes visualize the per-vertex body shape deviation from the target identity shape, where the color bar indicates the error range (0--50 mm). Diffusion-Based method better preserves the specified body shape while maintaining motion consistency with the action prompt.
    }
    \vspace{-10pt}
  \label{fig:quality_h3d}
\end{figure}

\noindent \textbf{Qualitative Results on HumanML3D.} We first provide a visual comparison on the HumanML3D test set to evaluate the motion-shape generation capabilities of different frameworks in Fig.~\ref{fig:quality_h3d}. Both our VQ-based and Diffusion-based models ($\imgf \iden$) directly synthesize body shape and motion sequences from the provided prompts. For the Shape My Moves baseline, which requires templated numeric measurements not present in the standard dataset, we use ground-truth $\beta$ values to retrieve the corresponding numeric template parameters. These retrieved values and action prompts are then fed into their pre-trained checkpoint to generate the results. Visualizations of per-vertex body shape deviation demonstrate that our diffusion-based model better preserves the target identity shape while maintaining motion consistency. While Shape My Moves can generate semantically plausible motions, the generated sequences tend to exhibit less accurate body-shape preservation and weaker fine-grained motion details.

\begin{figure}[h]
  \centering
  \includegraphics[width=\linewidth]{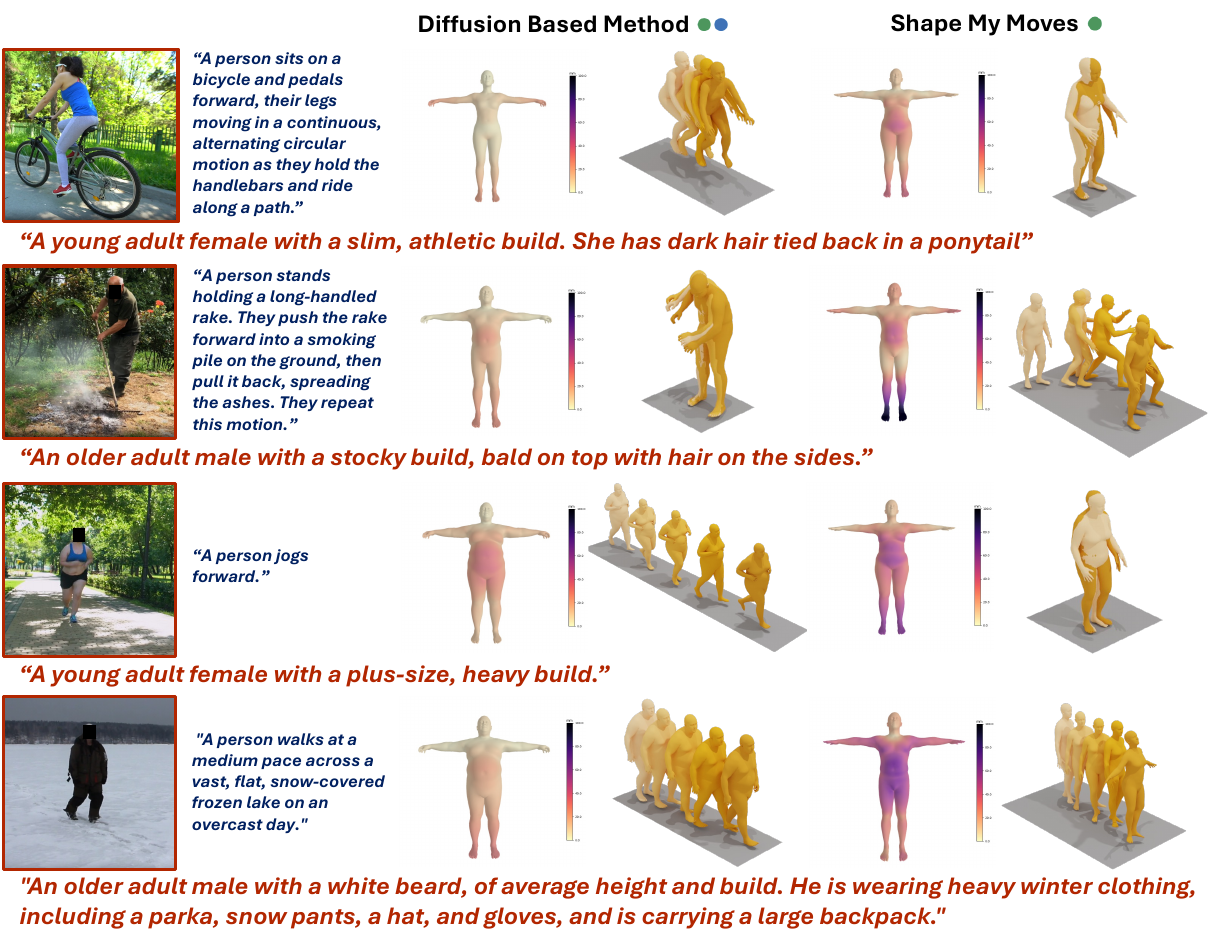}
  \caption{\textbf{Qualitative Results of Zero-shot Generalization on Unseen Test Set.}
  Each row shows an example consisting of a reference image, a motion prompt, and an identity description. 
  The colored meshes visualize the per-vertex body shape deviation from the target identity shape, where the color bar indicates the error magnitude in millimeters (0–100 mm). Our diffusion-based method better preserves the specified body shape while generating motions consistent with the action prompt.
  }
  \label{fig:quanlity_sstk}
\end{figure}

\noindent \textbf{Zero-shot Generalization on Unseen Set.} To assess the robustness of our model in real-world scenarios, we conduct zero-shot evaluations using the IdentityMotion test set, which contains identities strictly disjoint from the training data. Using our dual-conditioned diffusion model ($\imgf \iden$), we generate motions based on reference images, motion prompts, and identity descriptions. As illustrated in Fig.~\ref{fig:quanlity_sstk}, the model successfully synthesizes motions for a wide range of challenging identities. Despite the increased error range compared to HumanML3D test set, our method maintains strong motion-prompt adherence while preserving the target identity shape, whereas Shape My Moves frequently fails to follow the motion prompts.

\begin{figure}[t]
  \centering
  \includegraphics[width=1.0\linewidth]{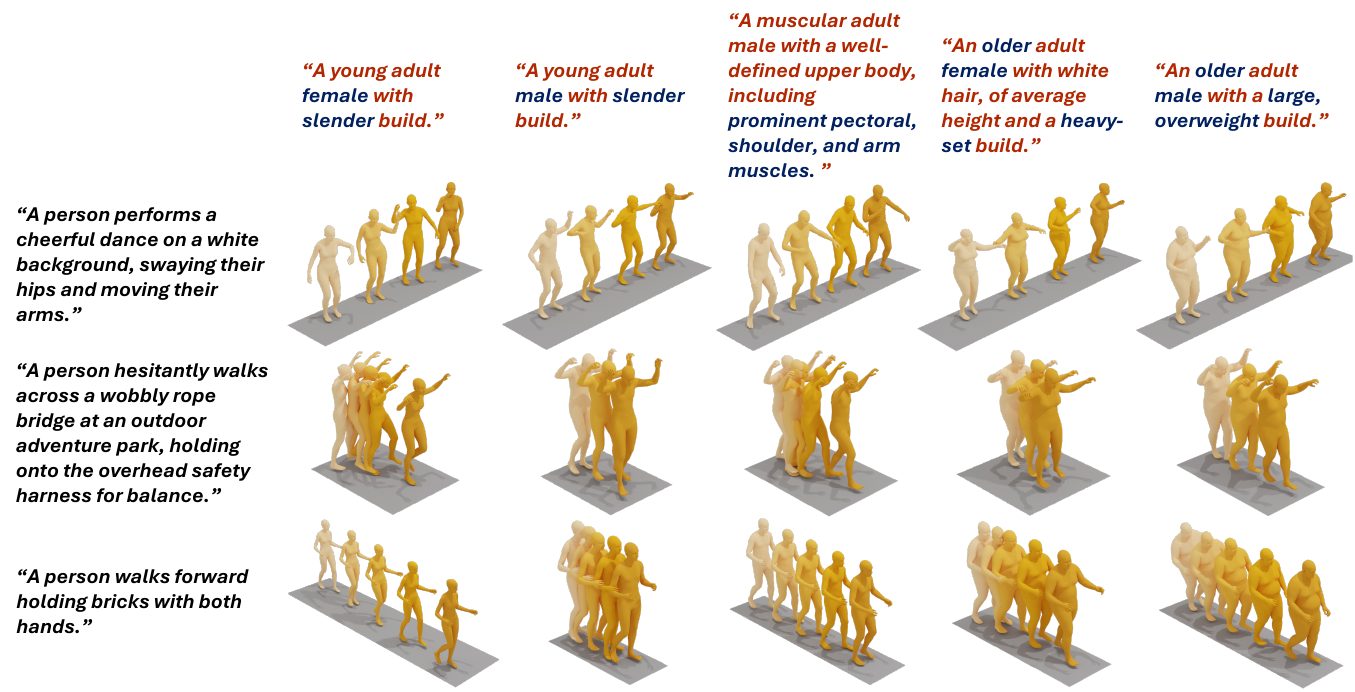}
  \caption{\textbf{Zero-shot Identity-Controllable Motion Generation.}
  Columns correspond to different identity descriptions, while rows correspond to different action prompts. The generated motion sequences follow the same action prompt while adapting motion style and body proportions to match the specified identity, demonstrating the model’s ability to disentangle identity from action. A small spatial offset is added to the first row to avoid occlusion.
  }
  \vspace{-10pt}
  \label{fig:ctrl_gen}
\end{figure}

\noindent \textbf{Identity-Controllable Motion Generation.} 
Finally, we evaluate whether the model can disentangle identity and motion by performing identity-controllable generation with randomly composed identity–motion prompts. Specifically, identity descriptions and motion prompts from the IdentityMotion dataset are randomly paired to form previously unseen combinations. In this setting, we utilize the dual-conditioned model ($\imgf \iden$) but zero-pad the image feature input, as reference images are not available for these identity–motion pairs. Fig.~\ref{fig:ctrl_gen} demonstrates that the model can successfully apply a single action prompt across a diverse spectrum of identities, ranging from ``slender'' to ``large, overweight'' builds. The generated motion sequences follow the same action logic while naturally adapting body proportions to match the specified descriptions. This ability to disentangle and independently control identity and motion underscores the model's capacity to learn a structured and versatile latent space. These qualitative results are further supported by a human perception study in our Supplementary, confirming our superiority in capturing identity-consistent motions.


\vspace{-10pt}
\section{Limitations}

Despite its plausible performance, our framework has several limitations. First, shape reconstruction is sensitive to loose clothing or occlusions in reference images, which can introduce noise into the predicted parameters. Second, zero-shot evaluations show that while motion consistency remains high, absolute error increases for extreme body types (e.g., exceptional height or mass) that lie outside the training distribution. Future work could explore more robust identity encoders or incorporate geometric constraints to further enhance mesh fidelity under these extreme conditions.

\section{Conclusion} 
\vspace{-10pt}
In this paper, we present \textbf{\textit{IAM}}, a novel framework for identity-aware human motion generation paradigms. Through extensive experiments on the HumanML3D and IdentityMotion datasets, we demonstrated that our diffusion-based model, particularly when conditioned on multi-modal identity descriptors ($\imgf\iden$), significantly outperforms baselines in preserving body shape precision and motion fidelity. Our model achieves state-of-the-art FID on HumanML3D and strong zero-shot generalization to unseen identities. Qualitative comparisons further confirm that our approach effectively disentangles identity from motion, allowing for fine-grained control over body proportions across diverse action sequences. Our results highlight the importance of explicitly modeling body morphology in motion generation and demonstrate that identity-aware motion synthesis is a promising direction towards more realistic and controllable human animation.

\bibliographystyle{splncs04}
\bibliography{main}

@String(CVPR  = {IEEE Conf. Comput. Vis. Pattern Recog.})

@String(TOG   = {ACM Trans. Graph.})

@String(CVPR  = {CVPR})

@String(TOG   = {ACM TOG})

@inproceedings{kim2025personabooth,
  title={Personabooth: Personalized text-to-motion generation},
  author={Kim, Boeun and Jeong, Hea In and Sung, JungHoon and Cheng, Yihua and Lee, Jeongmin and Chang, Ju Yong and Choi, Sang-Il and Choi, Younggeun and Shin, Saim and Kim, Jungho and others},
  booktitle={Proceedings of the IEEE/CVF Conference on Computer Vision and Pattern Recognition},
  pages={22756--22765},
  year={2025}
}

@article{wang2025generating,
  title={Generating Attribute-Aware Human Motions from Textual Prompt},
  author={Wang, Xinghan and Xu, Kun and Li, Fei and Sheng, Cao and Yu, Jiazhong and Mu, Yadong},
  journal={arXiv preprint arXiv:2506.21912},
  year={2025}
}

@article{bjorkstrand2025unconditional,
  title={Unconditional Human Motion and Shape Generation via Balanced Score-Based Diffusion},
  author={Bj{\"o}rkstrand, David and Wang, Tiesheng and Bretzner, Lars and Sullivan, Josephine},
  journal={arXiv preprint arXiv:2510.12537},
  year={2025}
}

@article{sawdayee2025dance,
  title={Dance like a chicken: Low-rank stylization for human motion diffusion},
  author={Sawdayee, Haim and Guo, Chuan and Tevet, Guy and Zhou, Bing and Wang, Jian and Bermano, Amit H},
  journal={arXiv preprint arXiv:2503.19557},
  year={2025}
}

@article{ho2022classifier,
  title={Classifier-free diffusion guidance},
  author={Ho, Jonathan and Salimans, Tim},
  journal={arXiv preprint arXiv:2207.12598},
  year={2022}
}

@article{SMPL:2015,
    author = {Loper, Matthew and Mahmood, Naureen and Romero, Javier and Pons-Moll, Gerard and Black, Michael J.},
    title = {{SMPL}: A Skinned Multi-Person Linear Model},
    journal = {ACM Transactions on Graphics, (Proc. SIGGRAPH Asia)},
    month = oct,
    number = {6},
    pages = {248:1--248:16},
    publisher = {ACM},
    volume = {34},
    year = {2015}
}

@inproceedings{SMPL-X:2019,
    title = {Expressive Body Capture: 3D Hands, Face, and Body from a Single Image},
    author = {Pavlakos, Georgios and Choutas, Vasileios and Ghorbani, Nima and Bolkart, Timo and Osman, Ahmed A. A. and Tzionas, Dimitrios and Black, Michael J.},
    booktitle = {Proceedings IEEE Conf. on Computer Vision and Pattern Recognition (CVPR)},
    year = {2019}
}

@inproceedings{radford2021learning,
  title={Learning transferable visual models from natural language supervision},
  author={Radford, Alec and Kim, Jong Wook and Hallacy, Chris and Ramesh, Aditya and Goh, Gabriel and Agarwal, Sandhini and Sastry, Girish and Askell, Amanda and Mishkin, Pamela and Clark, Jack and others},
  booktitle={International conference on machine learning},
  pages={8748--8763},
  year={2021},
  organization={PmLR}
}

@article{sanh2019distilbert,
  title={DistilBERT, a distilled version of BERT: smaller, faster, cheaper and lighter},
  author={Sanh, Victor and Debut, Lysandre and Chaumond, Julien and Wolf, Thomas},
  journal={arXiv preprint arXiv:1910.01108},
  year={2019}
}

@article{grattafiori2024llama,
  title={The llama 3 herd of models},
  author={Grattafiori, Aaron and Dubey, Abhimanyu and Jauhri, Abhinav and Pandey, Abhinav and Kadian, Abhishek and Al-Dahle, Ahmad and Letman, Aiesha and Mathur, Akhil and Schelten, Alan and Vaughan, Alex and others},
  journal={arXiv preprint arXiv:2407.21783},
  year={2024}
}

@article{streuber2016body,
  title={Body talk: Crowdshaping realistic 3D avatars with words},
  author={Streuber, Stephan and Quiros-Ramirez, M Alejandra and Hill, Matthew Q and Hahn, Carina A and Zuffi, Silvia and O'Toole, Alice and Black, Michael J},
  journal={ACM Transactions on Graphics (TOG)},
  volume={35},
  number={4},
  pages={1--14},
  year={2016},
  publisher={ACM New York, NY, USA}
}

@InProceedings{Guo_2022_CVPR,
    author    = {Guo, Chuan and Zou, Shihao and Zuo, Xinxin and Wang, Sen and Ji, Wei and Li, Xingyu and Cheng, Li},
    title     = {Generating Diverse and Natural 3D Human Motions From Text},
    booktitle = {Proceedings of the IEEE/CVF Conference on Computer Vision and Pattern Recognition (CVPR)},
    month     = {June},
    year      = {2022},
    pages     = {5152-5161}
}

@article{tevet2024closd,
  title={Closd: Closing the loop between simulation and diffusion for multi-task character control},
  author={Tevet, Guy and Raab, Sigal and Cohan, Setareh and Reda, Daniele and Luo, Zhengyi and Peng, Xue Bin and Bermano, Amit H and van de Panne, Michiel},
  journal={arXiv preprint arXiv:2410.03441},
  year={2024}
}

@inproceedings{xiao2025motionstreamer,
  title={Motionstreamer: Streaming motion generation via diffusion-based autoregressive model in causal latent space},
  author={Xiao, Lixing and Lu, Shunlin and Pi, Huaijin and Fan, Ke and Pan, Liang and Zhou, Yueer and Feng, Ziyong and Zhou, Xiaowei and Peng, Sida and Wang, Jingbo},
  booktitle={Proceedings of the IEEE/CVF International Conference on Computer Vision},
  pages={10086--10096},
  year={2025}
}

@inproceedings{choutas2022accurate,
  title={Accurate 3D body shape regression using metric and semantic attributes},
  author={Choutas, Vasileios and M{\"u}ller, Lea and Huang, Chun-Hao P and Tang, Siyu and Tzionas, Dimitrios and Black, Michael J},
  booktitle={Proceedings of the IEEE/CVF Conference on Computer Vision and Pattern Recognition},
  pages={2718--2728},
  year={2022}
}

@inproceedings{zhong2024smoodi,
  title={Smoodi: Stylized motion diffusion model},
  author={Zhong, Lei and Xie, Yiming and Jampani, Varun and Sun, Deqing and Jiang, Huaizu},
  booktitle={European Conference on Computer Vision},
  pages={405--421},
  year={2024},
  organization={Springer}
}

@inproceedings{guo2024momask,
  title={Momask: Generative masked modeling of 3d human motions},
  author={Guo, Chuan and Mu, Yuxuan and Javed, Muhammad Gohar and Wang, Sen and Cheng, Li},
  booktitle={Proceedings of the IEEE/CVF Conference on Computer Vision and Pattern Recognition},
  pages={1900--1910},
  year={2024}
}

@inproceedings{liao2025shape,
  title={Shape my moves: Text-driven shape-aware synthesis of human motions},
  author={Liao, Ting-Hsuan and Zhou, Yi and Shen, Yu and Huang, Chun-Hao Paul and Mitra, Saayan and Huang, Jia-Bin and Bhattacharya, Uttaran},
  booktitle={Proceedings of the Computer Vision and Pattern Recognition Conference},
  pages={1917--1928},
  year={2025}
}

@article{zhang2022motiondiffuse,
title   =   {MotionDiffuse: Text-Driven Human Motion Generation with Diffusion Model}, 
author  =   {Zhang, Mingyuan and Cai, Zhongang and Pan, Liang and Hong, Fangzhou and Guo, Xinying and Yang, Lei and Liu, Ziwei},
year    =   {2022},
journal =   {arXiv preprint arXiv:2208.15001},
}

@conference{AMASS:ICCV:2019,
  title = {{AMASS}: Archive of Motion Capture as Surface Shapes},
  author = {Mahmood, Naureen and Ghorbani, Nima and Troje, Nikolaus F. and Pons-Moll, Gerard and Black, Michael J.},
  booktitle = {International Conference on Computer Vision},
  pages = {5442--5451},
  month = oct,
  year = {2019},
  month_numeric = {10}
}

@InProceedings{tripathi2024humos,
    author    = {Tripathi, Shashank and Taheri, Omid and Lassner, Christoph and Black, Michael J. and Holden, Daniel and Stoll, Carsten},
    title     = {{HUMOS}: Human Motion Model Conditioned on Body Shape},
    booktitle = {European Conference on Computer Vision},
    organization = {Springer},
    year      = {2025},
    pages     = {133--152},
}

@misc{fan2025zerozeroshotmotiongeneration,
      title={Go to Zero: Towards Zero-shot Motion Generation with Million-scale Data}, 
      author={Ke Fan and Shunlin Lu and Minyue Dai and Runyi Yu and Lixing Xiao and Zhiyang Dou and Junting Dong and Lizhuang Ma and Jingbo Wang},
      year={2025},
      eprint={2507.07095},
      archivePrefix={arXiv},
      primaryClass={cs.CV},
      url={https://arxiv.org/abs/2507.07095}, 
}

@inproceedings{shen2024gvhmr,
  title={World-Grounded Human Motion Recovery via Gravity-View Coordinates},
  author={Shen, Zehong and Pi, Huaijin and Xia, Yan and Cen, Zhi and Peng, Sida and Hu, Zechen and Bao, Hujun and Hu, Ruizhen and Zhou, Xiaowei},
  booktitle={SIGGRAPH Asia Conference Proceedings},
  year={2024}
}

@article{comanici2025gemini,
  title={Gemini 2.5: Pushing the frontier with advanced reasoning, multimodality, long context, and next generation agentic capabilities},
  author={Comanici, Gheorghe and Bieber, Eric and Schaekermann, Mike and Pasupat, Ice and Sachdeva, Noveen and Dhillon, Inderjit and Blistein, Marcel and Ram, Ori and Zhang, Dan and Rosen, Evan and others},
  journal={arXiv preprint arXiv:2507.06261},
  year={2025}
}

@inproceedings{chen2023executing,
  title={Executing your commands via motion diffusion in latent space},
  author={Chen, Xin and Jiang, Biao and Liu, Wen and Huang, Zilong and Fu, Bin and Chen, Tao and Yu, Gang},
  booktitle={Proceedings of the IEEE/CVF conference on computer vision and pattern recognition},
  pages={18000--18010},
  year={2023}
}

@inproceedings{zhang2023generating,
  title={T2M-GPT: Generating Human Motion from Textual Descriptions with Discrete Representations},
  author={Zhang, Jianrong and Zhang, Yangsong and Cun, Xiaodong and Huang, Shaoli and Zhang, Yong and Zhao, Hongwei and Lu, Hongtao and Shen, Xi},
  booktitle={Proceedings of the IEEE/CVF Conference on Computer Vision and Pattern Recognition (CVPR)},
  year={2023},
}

@article{jiang2024motiongpt,
  title={Motiongpt: Human motion as a foreign language},
  author={Jiang, Biao and Chen, Xin and Liu, Wen and Yu, Jingyi and Yu, Gang and Chen, Tao},
  journal={Advances in Neural Information Processing Systems},
  volume={36},
  year={2024}
}

@inproceedings{fan2025go,
  title={Go to zero: Towards zero-shot motion generation with million-scale data},
  author={Fan, Ke and Lu, Shunlin and Dai, Minyue and Yu, Runyi and Xiao, Lixing and Dou, Zhiyang and Dong, Junting and Ma, Lizhuang and Wang, Jingbo},
  booktitle={Proceedings of the IEEE/CVF International Conference on Computer Vision},
  pages={13336--13348},
  year={2025}
}

@inproceedings{zhang2023skinned,
  title={Skinned motion retargeting with residual perception of motion semantics \& geometry},
  author={Zhang, Jiaxu and Weng, Junwu and Kang, Di and Zhao, Fang and Huang, Shaoli and Zhe, Xuefei and Bao, Linchao and Shan, Ying and Wang, Jue and Tu, Zhigang},
  booktitle={Proceedings of the IEEE/CVF Conference on Computer Vision and Pattern Recognition},
  pages={13864--13872},
  year={2023}
}

@article{jiang2019synthesis,
  title={Synthesis of biologically realistic human motion using joint torque actuation},
  author={Jiang, Yifeng and Van Wouwe, Tom and De Groote, Friedl and Liu, C Karen},
  journal={ACM Transactions On Graphics (TOG)},
  volume={38},
  number={4},
  pages={1--12},
  year={2019},
  publisher={ACM New York, NY, USA}
}

@inproceedings{zhang2024generative,
  title={Generative motion stylization of cross-structure characters within canonical motion space},
  author={Zhang, Jiaxu and Chen, Xin and Yu, Gang and Tu, Zhigang},
  booktitle={Proceedings of the 32nd ACM International Conference on Multimedia},
  pages={7018--7026},
  year={2024}
}

@article{zhang2023tapmo,
  title={Tapmo: Shape-aware motion generation of skeleton-free characters},
  author={Zhang, Jiaxu and Huang, Shaoli and Tu, Zhigang and Chen, Xin and Zhan, Xiaohang and Yu, Gang and Shan, Ying},
  journal={arXiv preprint arXiv:2310.12678},
  year={2023}
}

@article{shi2025motionpersona,
  title={MotionPersona: Characteristics-aware Locomotion Control},
  author={Shi, Mingyi and Liu, Wei and Mei, Jidong and Tse, Wangpok and Chen, Rui and Chen, Xuelin and Komura, Taku},
  journal={arXiv preprint arXiv:2506.00173},
  year={2025}
}

@article{Lee:2021:Parameterized,
  author = {Lee, Seyoung and Lee, Sunmin and Lee, Yongwoo and Lee, Jehee},
  title = {Learning a family of motor skills from a single motion clip},
  journal = {ACM Trans. Graph.},
  volume = {40},
  number = {4},
  year = {2021},
  articleno = {93},
}

@article{xu2023adaptnet,
  title={Adaptnet: Policy adaptation for physics-based character control},
  author={Xu, Pei and Xie, Kaixiang and Andrews, Sheldon and Kry, Paul G and Neff, Michael and McGuire, Morgan and Karamouzas, Ioannis and Zordan, Victor},
  journal={ACM Transactions on Graphics (TOG)},
  volume={42},
  number={6},
  pages={1--17},
  year={2023},
  publisher={ACM New York, NY, USA}
}

@inproceedings{jiang2025solami,
  title={Solami: Social vision-language-action modeling for immersive interaction with 3d autonomous characters},
  author={Jiang, Jianping and Xiao, Weiye and Lin, Zhengyu and Zhang, Huaizhong and Ren, Tianxiang and Gao, Yang and Lin, Zhiqian and Cai, Zhongang and Yang, Lei and Liu, Ziwei},
  booktitle={Proceedings of the Computer Vision and Pattern Recognition Conference},
  pages={26887--26898},
  year={2025}
}

@article{zhang2025vibes,
  title={ViBES: A Conversational Agent with Behaviorally-Intelligent 3D Virtual Body},
  author={Zhang, Juze and Chen, Changan and Chen, Xin and Yu, Heng and Xiang, Tiange and Khan, Ali Sartaz and Lakshmikanth, Shrinidhi K and Adeli, Ehsan},
  journal={arXiv preprint arXiv:2512.14234},
  year={2025}
}

@inproceedings{wu2025uniphys,
  title={Uniphys: Unified planner and controller with diffusion for flexible physics-based character control},
  author={Wu, Yan and Karunratanakul, Korrawe and Luo, Zhengyi and Tang, Siyu},
  booktitle={Proceedings of the IEEE/CVF International Conference on Computer Vision},
  pages={13214--13224},
  year={2025}
}

@article{xie2026textop,
  title={TextOp: Real-time Interactive Text-Driven Humanoid Robot Motion Generation and Control},
  author={Xie, Weiji and Zheng, Jiakun and Han, Jinrui and Shi, Jiyuan and Zhang, Weinan and Bai, Chenjia and Li, Xuelong},
  journal={arXiv preprint arXiv:2602.07439},
  year={2026}
}

@inproceedings{lu2025scamo,
  title={Scamo: Exploring the scaling law in autoregressive motion generation model},
  author={Lu, Shunlin and Wang, Jingbo and Lu, Zeyu and Chen, Ling-Hao and Dai, Wenxun and Dong, Junting and Dou, Zhiyang and Dai, Bo and Zhang, Ruimao},
  booktitle={Proceedings of the Computer Vision and Pattern Recognition Conference},
  pages={27872--27882},
  year={2025}
}

@article{li2026llamo,
      title={LLaMo: Scaling Pretrained Language Models for Unified Motion Understanding and Generation with Continuous Autoregressive Tokens}, 
      author={Zekun Li and Sizhe An and Chengcheng Tang and Chuan Guo and Ivan Shugurov and Linguang Zhang and Amy Zhao and Srinath Sridhar and Lingling Tao and Abhay Mittal},
      year={2026},
      journal={arXiv preprint arXiv:2602.12370}
}

@article{wen2025hy,
  title={HY-Motion 1.0: Scaling Flow Matching Models for Text-To-Motion Generation},
  author={Wen, Yuxin and Shuai, Qing and Kang, Di and Li, Jing and Wen, Cheng and Qian, Yue and Jiao, Ningxin and Chen, Changhai and Chen, Weijie and Wang, Yiran and others},
  journal={arXiv preprint arXiv:2512.23464},
  year={2025}
}

@inproceedings{zhong2023attt2m,
  title={Attt2m: Text-driven human motion generation with multi-perspective attention mechanism},
  author={Zhong, Chongyang and Hu, Lei and Zhang, Zihao and Xia, Shihong},
  booktitle={Proceedings of the IEEE/CVF international conference on computer vision},
  pages={509--519},
  year={2023}
}

@article{jiang2023motiongpt,
  title={Motiongpt: Human motion as a foreign language},
  author={Jiang, Biao and Chen, Xin and Liu, Wen and Yu, Jingyi and Yu, Gang and Chen, Tao},
  journal={Advances in Neural Information Processing Systems},
  volume={36},
  pages={20067--20079},
  year={2023}
}

@article{tevet2022human,
  title={Human motion diffusion model},
  author={Tevet, Guy and Raab, Sigal and Gordon, Brian and Shafir, Yonatan and Cohen-Or, Daniel and Bermano, Amit H},
  journal={arXiv preprint arXiv:2209.14916},
  year={2022}
}

@inproceedings{chen2025language,
  title={The language of motion: Unifying verbal and non-verbal language of 3d human motion},
  author={Chen, Changan and Zhang, Juze and Lakshmikanth, Shrinidhi K and Fang, Yusu and Shao, Ruizhi and Wetzstein, Gordon and Fei-Fei, Li and Adeli, Ehsan},
  booktitle={Proceedings of the Computer Vision and Pattern Recognition Conference},
  pages={6200--6211},
  year={2025}
}
\clearpage
\appendix

\section{Video Demonstration}
The supplementary video provides a comprehensive visualization of our work, including animated results for all figures in the main paper. We recommend viewing the video to better evaluate the execution and identity-aware motion dynamics of our method.

\begin{figure}[h]
  \centering
  \includegraphics[width=\linewidth]{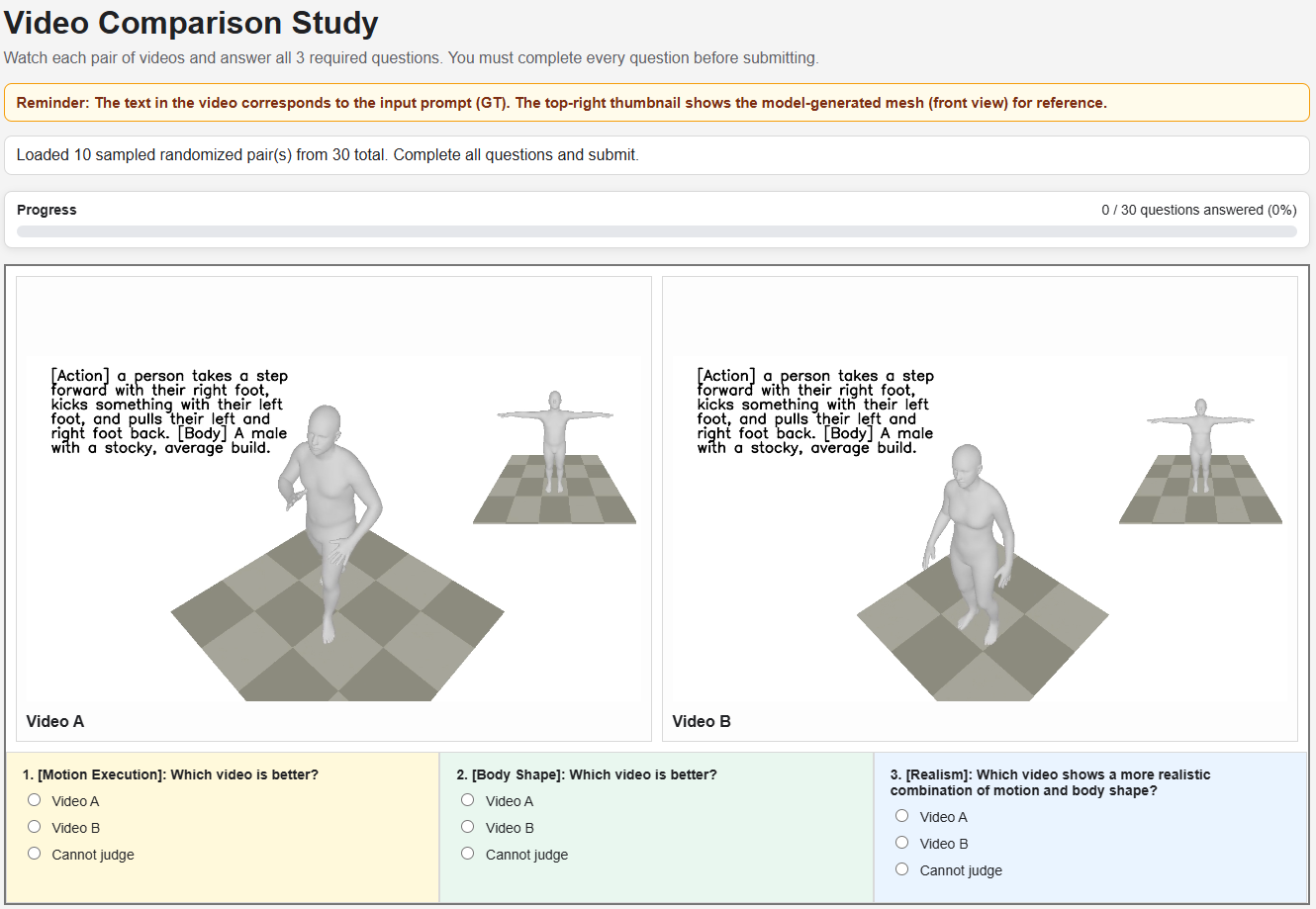}
  \caption{\textbf{User study interface.}
    Each trial presents two anonymous videos (A/B), the input prompt, and a frontal mesh reference. Participants select which video better matches motion, body shape, and overall motion--shape realism.
    }
  \label{fig:userstudy}
\end{figure}

\section{Human Perception Study}
To evaluate subjective quality, we conducted a perception study comparing our method against Shape My Move.
We collected 25 valid responses, where each participant evaluated 10 trials drawn from a pool of 30 randomly sampled prompt-video pairs from the HumanML3D test set. Our model was trained exclusively on HumanML3D to ensure a fair comparison with the baseline.

\noindent \textbf{Protocol.} As shown in Fig.~\ref{fig:userstudy}, participants performed side-by-side comparisons across three metrics: (1) \textit{Motion Plausibility} (alignment with the text prompt), (2) \textit{Shape Plausibility} (accuracy of body morphology), and (3) \textit{Motion-Shape Realism} (physical synergy between motion and shape). We put an extra forced-choice paradigm with a ``Cannot judge'' option to minimize bias.

\begin{figure}[h]
  \centering
  \includegraphics[width=0.5\linewidth]{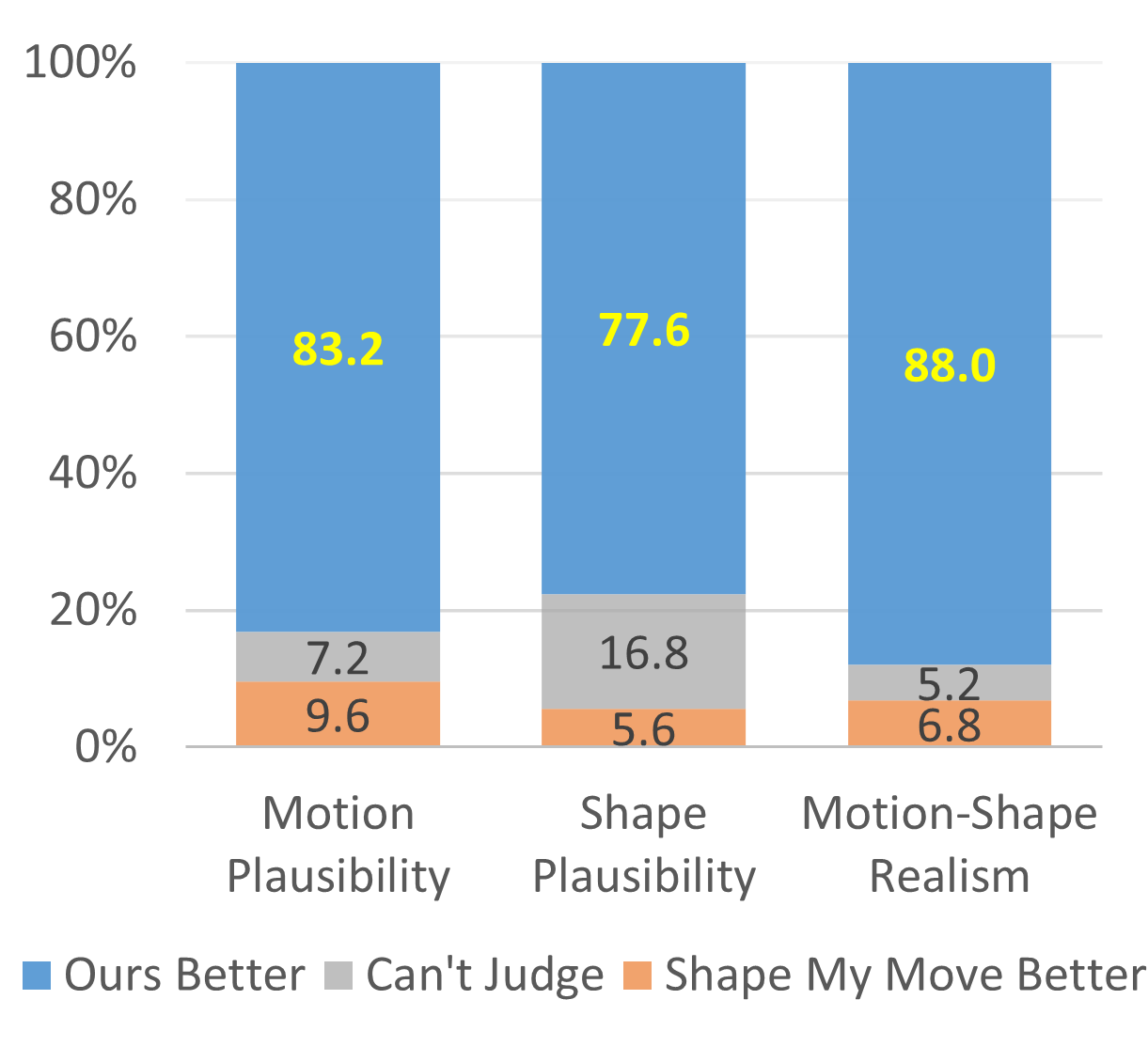}
  \caption{User Study Results.
    }
  \label{fig:userstudychart}
\end{figure}

\noindent \textbf{Results.} Quantitative results (Fig.~\ref{fig:userstudychart}) show that our method is significantly preferred across all criteria ($p < 0.05$ in all cases). Notably, our model demonstrates superior \textit{identity-motion synergy}, indicating a more physically plausible coupling between specific body builds and their corresponding action dynamics compared to the baseline.

\section{IdentityMotion Annotation Prompt}

\noindent \textbf{Gemini Annotation Prompt.} We provide the prompt used for data annotation with Gemini 2.5 Pro.

\begin{figure*}[t!]
\centering
\resizebox{0.99\textwidth}{!}{
\begin{planbox}{Gemini Annotation Prompt} 
\small
\textbf{Task Overview}\\
You will be shown a human motion video. Your goal is to analyze the motion in detail and then write several concise text prompts that could be used to generate this motion in a text-to-motion model.

Let's do this step by step.

\textbf{Step 1: Detailed Motion Analysis}\\
Fill in the following JSON fields with concise, accurate, and unambiguous information based on the video.

- body\_part\_involvement: Describe the main limbs and body parts involved in the movement (e.g., "Full body, with emphasis on legs and core").

- action\_sequence: A coherent, chronological description of the discrete actions. Focus on the 'what' (e.g., "The person takes a step forward with the right foot, then squats down, then pushes up into a jump").

- temporal\_progression: Describe how different body parts move in relation to one another over time. Focus on the 'how' and 'coordination' (e.g., "As the knees bend into the squat, the arms swing backward for momentum. The upward jump is initiated by a rapid extension of the legs, followed by the arms swinging forward and upward").

\textbf{Step 2: Physical Description}\\
Fill in the following JSON fields with concise, accurate, and unambiguous information based on the video.

- body\_description: Describe the person's apparent age, gender, and body build (e.g., height, proportion, muscularity).

\textbf{Step 3: Classification Checklist}\\
For each dimension below, provide the corresponding label based on the given video.
Choose \textbf{EXACTLY ONE} from the provided list and fill them in the following JSON fields.

- Age: `[Baby, Child, YoungAdult, Adult, MiddleAged, OlderAdult]`

- Gender: `[Male, Female, Unspecified]`

- ActionType: `[Locomotion, Manipulation, Exercise, Performance, Recreation]`

- Scene: `[Indoor, Outdoor]`

\textbf{Optional fields}\\
- movement\_style: e.g., smooth, explosive, robotic, fluid, hesitant.

- motion\_speed: Overall motion speed or intensity (low, medium, high).

- slow\_motion: If it is a slow motion video or not (true, false).

- static: If the human motion and video content is almost static or not (true, false).

- detailed\_action: Very detailed action description in a few words (e.g., "ballet dancing" instead of "dancing", "serving in tennis" instaed of "tennis", "swimming butterfly stroke" instead of "swimming").

\textbf{Step 4: Final Prompt Synthesis}\\
Actively use the information you identified in previous analysis from Step 1, Step2, Step3, and the [Optional fields], generate a list of 5 distinct text prompts with diverse and detailed prompts.

- The goal is to create prompts with different levels of conditioning information. For example:\\
  - (Base Prompt) A simple, high-level description of the core action.\\
    - e.g., "A man does a lunge forward."\\
  - (Styled Prompt) Incorporate the `movement\_style`.\\
    - e.g., "A person performs a slow and controlled lunge."\\
  - (Motion Script) A clear, sequential description of the actions, written in the present tense, much like a script's action lines.\\
    - e.g., "The young woman raises their hand to wave, takes one step to their right, and then extends their arm to point forward."\\
  - (Detailed Prompt) Combine multiple elements for a rich description.\\
    - e.g., "An athletic man performs an explosive lunge forward on a running track."\\
\end{planbox}
}
\vspace{-0.2cm}
\label{fig:gemini_annot_1}
\end{figure*}

\begin{figure*}[t!]
\centering
\resizebox{0.99\textwidth}{!}{
\begin{planbox}{Gemini Annotation Prompt Cont.} 
\small
\textbf{Clarity Rules}
- Be specific. Instead of "moves arm up," use "extends right arm forward until it is parallel with the ground."\\
- All directional descriptions (left, right, forward, back) must be from the perspective of the person performing the action.\\
- If a person is stationary, state this explicitly and decribie it more like a poseture (e.g., "The person stands still,").

\hrulefill

\textbf{Output format}
Return only a valid JSON object. Do not output any text outside of the JSON object. In the 'motion\_prompt' field, return a list of 5 distinct ONLY text prompts separated by comma, DON'T include \textbf{Base Prompt:} or \textbf{Styled Prompt:} such type information.

\{\\
\quad \quad "body\_part\_involvement": "...",\\
\quad \quad "action\_sequence": "...",\\
\quad \quad "temporal\_progression": "...",\\
\quad \quad "body\_description": "...",\\
\quad \quad "classification": \{\\
\quad \quad \quad "age": "...",\\
\quad \quad \quad "gender": "...",\\
\quad \quad \quad "action\_type": "...",\\
\quad \quad \quad "scene": "..."\\
\quad \quad \},\\
\quad \quad "motion\_prompt": [\\
\quad \quad \quad "...",\\
\quad \quad \quad "...",\\
\quad \quad \quad "..."\\
\quad \quad ],\\
\quad \quad "movement\_style": "...",\\
\quad \quad "motion\_speed": "...",\\
\quad \quad "slow\_motion": "...",\\
\quad \quad "static": "...",\\
\quad \quad "detailed\_action": "..."\\
\}

\end{planbox}
}
\vspace{-0.2cm}
\label{fig:gemini_annot_2}
\end{figure*}

\noindent \textbf{Llama 3.2 Neutralization Prompt.} We also provide the prompt utilized to anonymize identity-related descriptors from the initial Gemini-generated annotations, leveraging the Llama 3.2 model.
\begin{figure*}[t!]
\centering
\resizebox{0.99\textwidth}{!}{
\begin{planbox}{Llama Neutralization Prompt} 
\small
\textbf{Task}\\
You will be given a list of motion description of a human performing an action.

Your task is to neutralize the description by removing any identity-related, role-related, scene-related, or body-build information, while keeping all motion details intact.

\hrulefill

\textbf{Rules}\\
- \textbf{CRITICAL: STRICTLY MUST} replace all gendered pronouns (he, she, him, her, his, hers) with neutral forms ("they", "them", or "their"). This is mandatory for every sentence.\\
- Replace words describing \textbf{identity or gender} (e.g., man, woman, boy, girl, male, female)\\
  → Replace with neutral human references such as "a person", "an individual", or "a human".\\
  → Keep grammatical consistency (e.g., "A man" => "A person", "An old woman" => "An individual").\\
- Remove words describing \textbf{age, body build, or appearance}
  (e.g., young, old, adult, fit, slender, muscular, heavyset, tall, short, beautiful, handsome, etc.)\\
- Remove all \textbf{garment descriptions}
  (e.g., blue T-shirt, shorts, white dress, red swimsuit).\\
- Remove \textbf{occupational or role terms} implying identity
  (e.g., dancer, athlete, martial artist, performer, worker, actor).\\
- Keep all motion details (actions, postures, transitions, tempo, limb movements).\\
- Do not infer or add new details.\\
- \textbf{CRITICAL: STRICTLY MUST} rewrite the sentence so that it is grammatically correct and fluent.\\
- \textbf{CRITICAL: STRICTLY MUST} process EACH element independently (do not merge or deduplicate them).

\hrulefill

\textbf{Output format}
Output the rewritten, neutralized motion descriptions in the same list format as the input, e.g.:

"neutralized\_prompt": [\\
\quad \quad  "...", \\
\quad \quad  "...", \\
\quad \quad  "..." \\
]

Do not print any commentary or code.
\end{planbox}
}
\vspace{-0.2cm}
\label{fig:llama_annot}
\end{figure*}

\end{document}